\documentclass[letterpaper, 10 pt, conference]{ieeeconf}  % Comment this line out if you need a4paper

\IEEEoverridecommandlockouts                              % This command is only needed if 
\usepackage[T1]{fontenc}
\usepackage{times}
\usepackage{soul}
\usepackage{url}
\usepackage[hidelinks]{hyperref}
\usepackage{graphicx}
\usepackage{amsmath}
\usepackage{amssymb}
\usepackage{booktabs}
\usepackage{algorithm}
\usepackage{algorithmic}
\usepackage{multirow}
\usepackage{array}
\usepackage{xcolor}
\usepackage{cite}
\usepackage{float}
\definecolor{purple}{RGB}{153, 0, 255}

\usepackage{hhline}
\usepackage[caption=false]{subfig}
\usepackage{listings}
\usepackage{fancyvrb}
\usepackage{xcolor}
\definecolor{highlight}{HTML}{DAE8FC}
\definecolor{blue}{HTML}{4a86e8}
\definecolor{yellow}{HTML}{e69138}
\definecolor{pink}{HTML}{e06666}
\newcommand{\method}[0]{LLM-GROP}

\newcommand{\random}{TPRA}
\newcommand{\LLM}{LATP}
\usepackage{threeparttable}
\usepackage{caption}
\title{\LARGE \bf
Task and Motion Planning with Large Language Models\\ for Object Rearrangement
}

\author{Yan Ding$^{1*}$, Xiaohan Zhang$^{1*}$, Chris Paxton$^2$, Shiqi Zhang$^1$% <-this % stops a space
\thanks{$^*$ Equal Contribution}% <-this % stops a space
\thanks{$^1$~Department of Computer Science, The State University of New York at Binghamton \texttt{\{yding25; xzhan244; zhangs\}@binghamton.edu}}
\thanks{$^2$~Meta AI \texttt{cpaxton@meta.com}}%
}

\begin{document}

\maketitle
\thispagestyle{empty}
\pagestyle{empty}

%%%%%%%%%%%%%%%%%%%%%%%%%%%%%%%%%%%%%%%%%%%%%%%%%%%%%%%%%%%%%%%%%%%%%%%%%%%%%%%%
\begin{abstract}
%The paper presents a new approach for arranging objects into functional configurations, which is a crucial task for service robots.
Multi-object rearrangement is a crucial skill for service robots, and commonsense reasoning is frequently needed in this process. However, achieving commonsense arrangements requires knowledge about objects, which is hard to transfer to robots. 
Large language models (LLMs) are one potential source of this knowledge, but they do not naively capture information about plausible physical arrangements of the world. We propose \method{}, which uses prompting to extract commonsense knowledge about semantically valid object configurations from an LLM and instantiates them with a task and motion planner in order to generalize to varying scene geometry. %, in order to 
\method{} allows us to go from natural-language commands to human-aligned object rearrangement in varied environments.
Based on human evaluations, our approach achieves the highest rating while outperforming competitive baselines in terms of success rate while maintaining comparable cumulative action costs. 
Finally, we demonstrate a practical implementation of \method{} on a mobile manipulator in real-world scenarios.
Supplementary materials are available at:
\url{https://sites.google.com/view/llm-grop}
\end{abstract}

\section{Introduction}

\begin{figure}
% https://docs.google.com/drawings/d/1Gt-ZEinp6MoALrK4TaxO2ieBLFtH3kBpbVPdUftFri8/edit?usp=sharing
\begin{center}
    \includegraphics[width=0.47\textwidth]{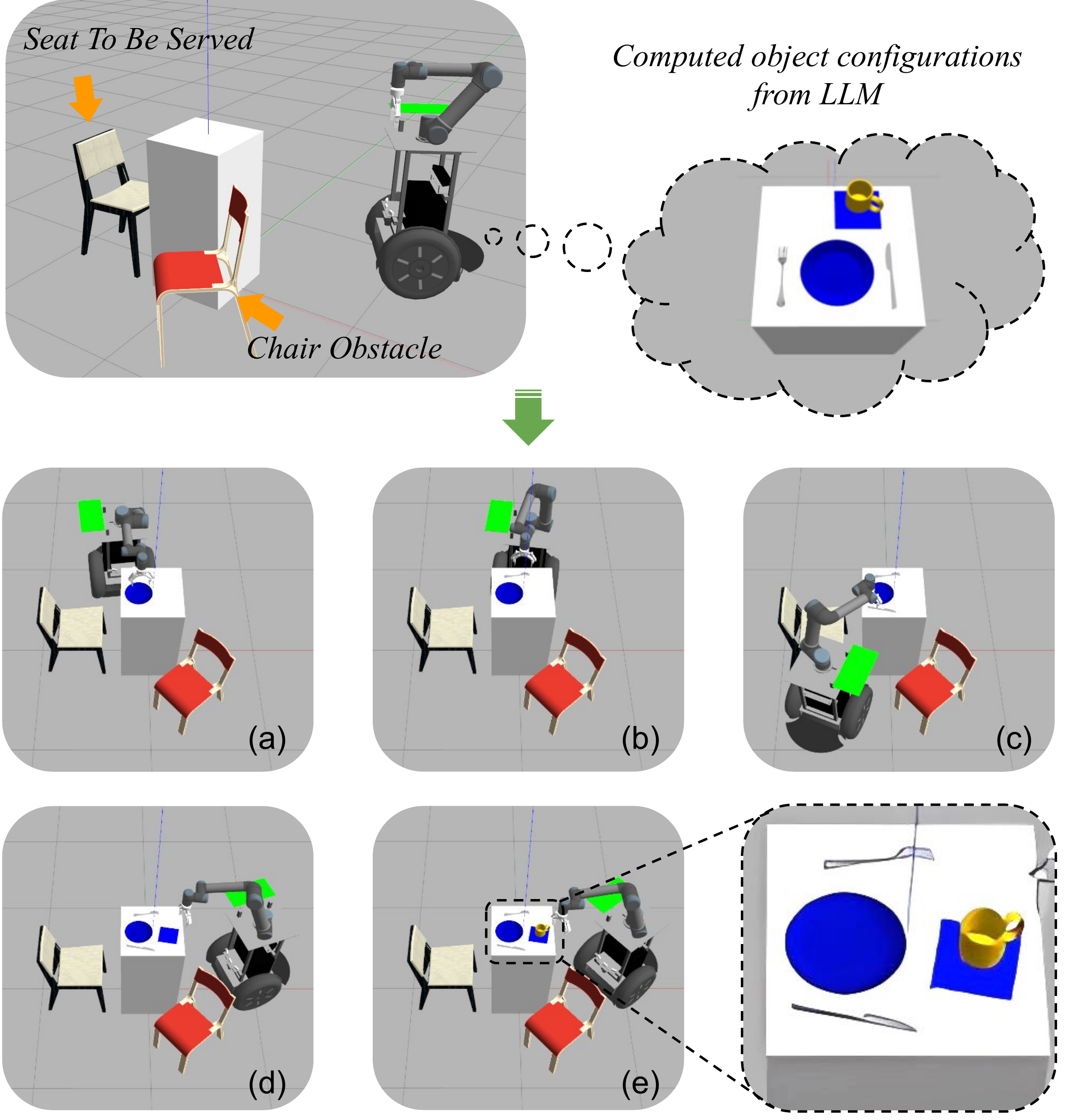}
    \vspace{-0.4em}
    \caption{A mobile manipulator is assigned the task of setting a table in a dining domain.
    The manipulator needs to arrange several tableware objects, including a knife, a fork, a plate, a cup mat, and a mug. 
    These objects are available on the other tables, and there are also randomly generated obstacles (i.e., the red chair) that are not included in the pre-built map beforehand.
    The robot needs to compute feasible and efficient plans for rearranging the objects on the target table using both navigation and manipulation behaviors.
    }
    \label{fig:set_table}
\end{center}
\vspace{-1.5em}
\end{figure}

Multi-object rearrangement is a critical skill for service robots to complete everyday tasks, such as setting tables, organizing bookshelves, and loading dishwashers~\cite{habitatrearrangechallenge2022,RoomR}.
These tasks demand robots exhibit both manipulation and navigation capabilities.
For example, a robot tasked with setting a dinner table might need to retrieve tableware objects like a fork or a knife from different locations and place them onto a table surrounded by chairs, as shown in Fig.~\ref{fig:set_table}.
To complete the task, the robot needs to correctly position the tableware objects in 
% \cpax{semantically meaningful not functional - fork left of right is not functional} 
semantically meaningful configurations (e.g., a fork is typically on the left of a knife) and efficiently navigate indoors while avoiding obstacles like chairs or humans whose locations are unknown in advance. 
% \cpax{chair isn't exactly dynamic - we just mean obstacles at unknown locations right?}
% \yan{Yes. The "chair" is not on the pre-built map. Here, I added one sentence "whose locations may not be known in advance." What do you think?}

% Mobile manipulation techniques for object rearrangement have made significant progress, as evidenced by studies
A variety of mobile manipulation systems have been developed for object rearrangement tasks~\cite{goodwin2022semantically,liu2022structformer,wei2023lego,huang2019large,gu2022multi,king2016rearrangement, cheong2020relocate,vasilopoulos2021reactive}.
% These approaches can be divided into two categories based on their reliance on common sense.
% The first category requires explicit instructions, such as arranging similar colored items in a line or placing them in a specific shape on a table.
% Examples of such studies include~\cite{goodwin2022semantically,huang2019large,gu2022multi,cheong2020relocate,vasilopoulos2021reactive,liang2022code}.
Most of those systems require explicit instructions, such as arranging similar colored items in a line or placing them in a specific shape on a table~\cite{goodwin2022semantically,huang2019large,gu2022multi,cheong2020relocate,vasilopoulos2021reactive,liang2022code}.
% The second category relies on a degree of common sense to arrange items under vague instructions, e.g., setting the table and rearranging furniture. 
% For instance, when setting a dinner table, it is customary to place forks on the right side of the plate and knives on the left. 
However, user requests in the real world tend to be underspecified: there can be many different ways to set a table that are not equally preferred. 
% \cpax{Removed a lot of "e.g." - feel free to revert, but in my opinion it breaks up the flow of the writing.}
How does a robot figure out a fork should be placed on the left of a plate and a knife on the right?
Considerable commonsense knowledge is needed. 
% \cpax{I would say here something like "recent results have shown large language models (LLMs) like GPT3 capture a great deal of this common sense knowledge~\cite{}"}.
Recent results have shown large language models (LLMs) like GPT3~\cite{brown2020language} and ChatGPT~\cite{openai} capture a great deal of this common sense knowledge~\cite{liu2021pre}.
% The TV cabinet is usually facing the sofa.
% Studies that fall under this category include~\cite{liu2022structformer,kapelyukh2022dall, liu2022structdiffusion, wei2023lego}. 
% However, acquiring common sense in these types of work typically requires extensive learning from demonstrations, which can be a costly and time-consuming process. 
% Additionally, scalability is a critical issue that cannot be overlooked.
In the past, researchers have equipped mobile manipulators with semantic information using machine learning methods~\cite{liu2022structformer,liu2022structdiffusion,wei2023lego,zhang2021hierarchical}. 
Those methods require collecting training data, which limits their applicability to robots working on complex service tasks.

To equip robot planning methods with common sense for object rearrangement, we introduce \method{}, standing for
% (\textbf{L}arge \textbf{L}anguage \textbf{M}odel-\textbf{G}uided \textbf{RO}bot Task and Motion \textbf{P}lanning), 
\textbf{L}arge \textbf{L}anguage \textbf{M}odel for \textbf{G}rounded \textbf{RO}bot Task and Motion \textbf{P}lanning, 
our approach that leverages commonsense knowledge for planning to complete object rearrangement tasks. 
% Initially, LLMs generate {symbolic spatial relationships} between objects, which are checked for logical consistency by a symbolic reasoning module.
LLM-GROP first uses an LLM to generate \emph{symbolic} spatial relationships between objects, e.g., a fork and a knife are placed on the left and right respectively. 
% , called Logic-based Conflict Detection in the task planner.
% Once a feasible spatial relationship is determined, LLMs proceed to compute {geometric spatial relationships} between objects, and an object-grounded adaptive sampler in the motion planner samples feasible object positions.
The spatial relationships then can be grounded to different \emph{geometric} spatial relationships whose feasibility levels are evaluated by a motion planning system, e.g., placing objects in some areas of a table can be easier than the others. 
Finally, the feasibility and efficiency of different task-motion plans  are optimized towards maximizing long-term utility, i.e., seeking the best trade-off between motion feasibility and task-completion efficiency. 

We have applied \method{} to a dining 
% \cpax{room?}
room, where a mobile manipulator must set a table according to a user's instructions. 
% through the random selection of utensils and their corresponding positions for assessment.
A set of tableware objects are provided to the robot, where the robot's task is to compute a tabletop configuration of those objects that comply with common sense, and compute a task-motion plan to realize the configuration. 
%by randomly sampling utensils and their positions for evaluation. 
%However, evaluating the success or failure of the rearrangement results is difficult as there are no fixed rules to determine this. 
% However, determining the success or failure of the rearrangement results posed a challenge since there are no established criteria to guide the evaluation. 
% It was a non-trivial task of evaluating the ``accordance'' of an arrangement with respect to commonsense rules, where human participants helped with subjective evaluations. 
To evaluate the performance of our approach, we had users rate different place settings to get a subjective evaluation.
% \cpax{this sentence is a bit unclear to me}
% \cpax{To evaluate the performance of our approach, we had users rate different place settings to get a subjective evaluation.}
% In response, we engaged \textcolor{red}{X} volunteers to evaluate the table-setting outcome, resulting in an overall rating of \textcolor{red}{Y} out of a possible 5.0 - the highest achievable score.
%To address this, we enlisted \textcolor{red}{X} volunteers to evaluate the table-setting outcome, which resulted in an overall rating of \textcolor{red}{Y} out of 5.0, the highest score achievable. 
%Our \method{} system produced a higher satisfactory rate compared to baselines, while also maintaining or reducing cumulative action costs. 
% Our \method{} system delivered a higher satisfaction rate compared to baselines while simultaneously maintaining or reducing cumulative action costs. 
% From the results, w
We observed improvements in user satisfaction from \method{} compared with existing object rearrangement methods, while maintaining similar or lower cumulative action costs. 
% Lastly, we demonstrated the practical implementation of \method{} by using actual robot hardware in a real-world scenario.
Finally, \method{} was demonstrated on a real robot.

\section{Related Work}
\label{sec:related}

% In this section, w
% We first introduce the object rearrangement domain, where tabletop object arrangement was largely overlooked. 
We first introduce the object rearrangement domain, then 
 % and object sorting
% \cpax{This seems like a strong statement! Maybe "in favor of..." and cite various other rearrangement problems?}
% \tbd_yan{I rephrased this sentence. What do you think is the current one? If you are ok with this statement, you can remove these comments.}
discuss methods for tabletop object arrangement that mostly rely on supervised learning methods, and finally summarize research on using large language models for planning.

\subsection{Object Rearrangement}
Rearranging objects is a critical task for service robots, and much research has focused on moving objects from one location to another and placing them in a new position. 
Examples include the Habitat Rearrangement Challenge~\cite{habitatrearrangechallenge2022} and the AI2-THOR Rearrangement Challenge~\cite{RoomR}.
There is rich literature on object rearrangement in robotics~\cite{goodwin2022semantically,huang2019large,gu2022multi,cheong2020relocate,vasilopoulos2021reactive,liang2022code,zhang2022visually}. 
A common assumption in those methods is that a goal arrangement is part of the input, and the robot knows the exact desired positions of objects. 
ALFRED~\cite{shridhar2020alfred} proposed a language-based multi-step object rearrangement task, for which a number of solutions have been proposed that combine high-level skills~\cite{blukis2022persistent,min2021film}, and which have recently been extended to use LLMs as input~\cite{inoue2022prompter}. However, these operate at a very coarse, discrete level, instead of making motion-level and placement decisions, and thus can't make granular decisions about common-sense object arrangements.
% \cpax{ALFRED~\cite{shridhar2020alfred} proposed a language-based multi-step object rearrangement task, for which a number of solutions have been proposed that combine high-level skills~\cite{blukis2022persistent,min2021film}, and which have recently been extended to use LLMs as input~\cite{inoue2022prompter}. However, these operate at a very coarse, discrete level, instead of making motion-level and placement decisions, and thus can't make granular decisions about common-sense object arrangements.}

% arranging items of a similar color in a line, and creating a specific shape on a table.
% Significant progress has been observed in mobile manipulation research, concentrating on following explicit instructions to place objects in a particular position~\cite{goodwin2022semantically,huang2019large,gu2022multi,cheong2020relocate,vasilopoulos2021reactive,liang2022code, zhang2022visually}, e.g., arranging items of a similar color in a line, and creating a specific shape on a table.
% These studies aim to successfully place objects on a tabletop, with Gu et al.~\cite{gu2022multi} proposing M3, a novel approach for addressing long-horizon mobile manipulation tasks in object rearrangement, utilizing manipulation and navigation skills with multiple-point goals to improve the success rate of object placement. 
By contrast, our work accepts underspecified instructions from humans, such as setting a dinner table with a few provided tableware objects.
\method{} has the capability to do common sense object rearrangement by extracting knowledge from LLMs, and operates both on a high level and on making motion-level placement decisions.
% \cpax{and operates both on a high level and on making motion-level placement decisions.}

\subsection{Predicting Complex Object Arrangements}
Object arrangement is a task that involves arranging items on a tabletop to achieve a specific functional, semantically valid goal configuration. 
This task requires not only the calculation of object positions but also adherence to common sense, such as placing forks to the left and knives to the right when setting a table. 
Previous studies in this area, such as~\cite{liu2022structformer,kapelyukh2022dall, liu2022structdiffusion, wei2023lego}, focused on predicting complex object arrangements based on vague instructions. 
For instance, StructFormer~\cite{liu2021structformer} is a transformer-based neural network for arranging objects into semantically meaningful structures based on natural-language instructions. 
% Another example is StructDiffusion~\cite{liu2022structdiffusion}, which combines a diffusion model and an object-centric transformer to build semantically meaningful structures in human environments without step-by-step instructions.
By comparison, our approach \method{} utilizes an LLM for commonsense acquisition to avoid the need of demonstration data for computing object positions. 
Additionally, we optimize the feasibility and efficiency of plans for placing tableware objects. 

There is recent research for predicting complex object arrangement using web-scale diffusion models~\cite{kapelyukh2022dall}. 
% One approach that utilizes pre-trained models to compute object positions is DALL-E-Bot~\cite{kapelyukh2022dall}. 
Their approach, called DALL-E-Bot, enables a robot to generate images based on a text description using DALL-E~\cite{ramesh2022hierarchical}, and accordingly arrange objects in a tabletop scenario. 
Similar to DALL-E-Bot, \method{} achieves zero-shot performance using pre-trained models, but it is not restricted to a single top-down view of a table.
% \cpax{but it is not restricted to a single top-down view of a table.}
In addition, we consider the uncertainty in manipulation and navigation, and optimize efficiency and feasibility in planning.
% While DALL-E-Bot achieves zero-shot performance using DALL-E without additional training or data collection, our approach takes a different paradigm and considers practicality and efficiency.

\subsection{Robot Planning with Large Language Models}
% task planning with large language models: housekeep, zero-shot planner, SayCan, inner monologue, PROGPROMPT
% motion planning with large language models: Code-as-Policies~\cite{liang2022code}.
%A very recent review article~\cite{openai} explored the i
Many LLMs have been developed in recent years, such as BERT~\cite{devlin2018bert}, GPT-3~\cite{brown2020language}, ChatGPT~\cite{openai}, CodeX~\cite{chen2021evaluating}, and OPT~\cite{zhang2022opt}. 
% \cpax{some of these references should also appear in intro imo}
These LLMs can encode a large amount of common sense~\cite{liu2021pre} and have been applied to robot task planning~\cite{kant2022housekeep,huang2022language,ahn2022can,huang2022inner,singh2022progprompt,ding2023integrating,liu2023llm+,zhao2023large,liu2022structdiffusion,wu2023tidybot,rana2023sayplan}. 
For instance, the work of Huang et. al. showed that LLMs can be used for task planning in household domains by iteratively augmenting prompts~\cite{huang2022language}. 
SayCan is another approach that enabled robot planning with affordance functions to account for action feasibility, where the service requests are specified in natural language (e.g., ``make breakfast'')~\cite{ahn2022can}. 
Compared with those methods, LLM-GROP optimizes both feasibility and efficiency while computing semantically valid geometric configurations. 

\section{The LLM-GROP Approach}
% In this work, we will explore the mobile manipulation domain, with a particular focus on the task of ``table setting''.
The objective of this task is to rearrange multiple tableware objects, 
% referred to as $o^i$ where $0\leq i\leq N$, 
which are initially scattered at different locations, into a tabletop configuration that is semantically valid and aligns with common sense.
% We assume that we already know the attributes, such as sizes and shapes, of the objects and the table, including those with different properties.
The robot is provided with prior knowledge about table shapes and locations, and equipped with skills of loading and unloading tableware objects. 
% There would be dynamic obstacles, such as chairs surrounding the table, which cannot be identified in a pre-created map.
There are dynamic obstacles, e.g., chairs around tables, that can only be sensed at planning time. 
% This creates a challenge for the robot to perform the table-setting task effectively.
We consider uncertainty in navigation and manipulation behaviors. 
For instance, the robot can fail in navigation  (at planning or execution time) when its goal is too close to tables or chairs, and it can fail in manipulation when it is not close enough to the target position. 
Note that uncertainties are treated as black boxes in this work.

\begin{figure}
\begin{center}
    \includegraphics[width=0.51\textwidth]{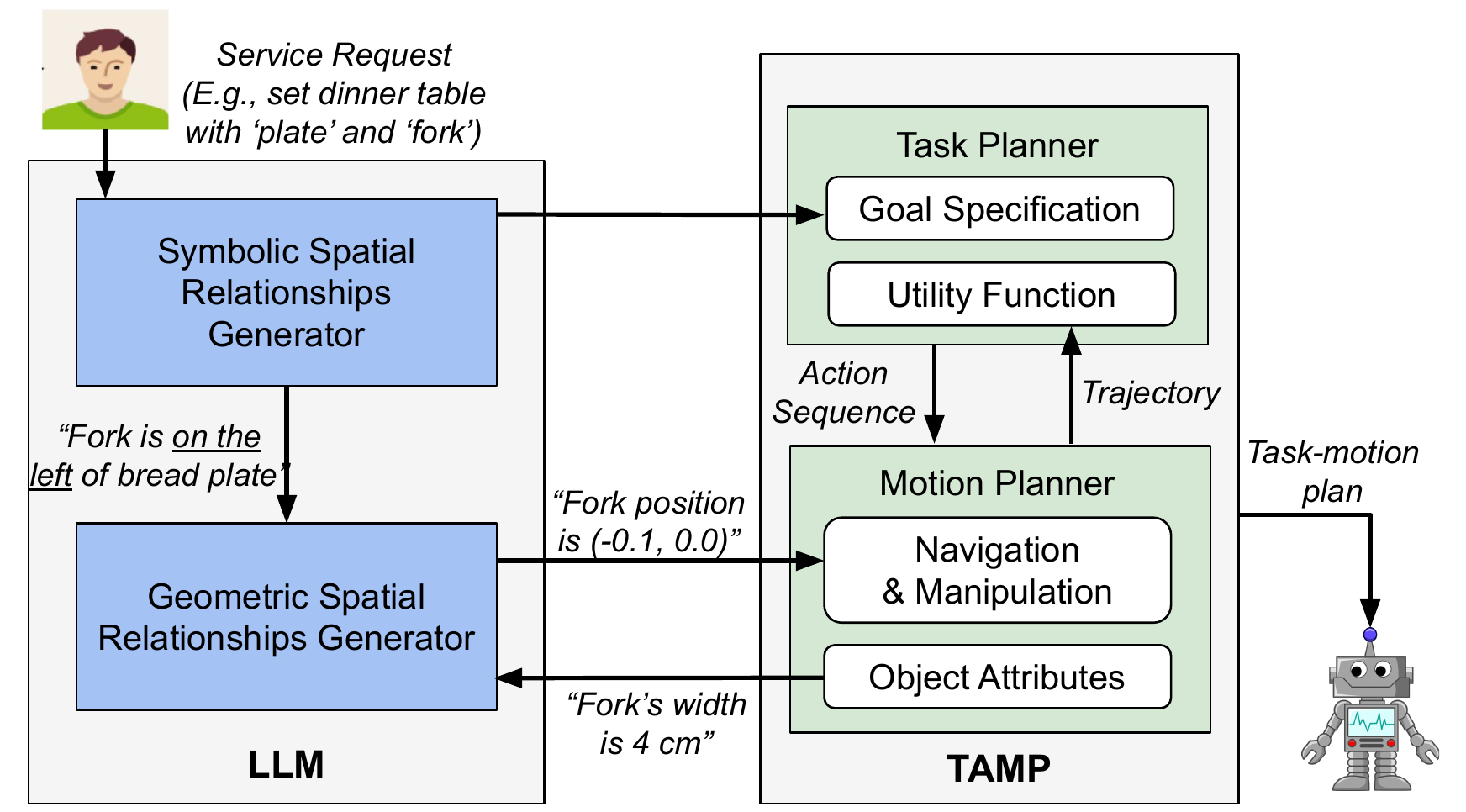} \vspace{-1.em}
    % \caption{The \method{} framework consists of essential components: {LLM} and {Task and Motion Planner}. 
    % The \textbf{\textcolor{blue}{blue}} loop represents the process of generating and examining spatial relationships between objects using {LLM} and {Logic-based Conflict Detection}.
    % The \textbf{\textcolor{yellow}{yellow}} lines indicate the adaptive sampling of object positions, where {LLM} computes objects' physical distance, and the motion planner employs {Object-Grounded Adaptive Sampler} to generate a few feasible object positions. 
    % These positions are evaluated in terms of feasibility and efficiency using a {utility function}.
    % The \textbf{\textcolor{pink}{pink}} loop illustrates the selection of the optimal object positions to maximize long-term utility.}
    \caption{
    \method{} takes service requests from humans for setting tables and produces a task-motion plan that the robot can execute. 
    \method{} is comprised of two key components: the LLM and the Task and Motion Planner.
    The LLM is responsible for creating both symbolic and geometric spatial relationships between the tableware objects. 
    This provides the necessary context for the robot to understand how the objects should be arranged on the table.
    The Task and Motion Planner generates the optimal plan for the robot to execute based on the information provided by the LLM.
    % \textcolor{red}{SHIQI: There should be an arrow pointing to ``Goal Specification''. Label the gray box with ``TAMP''. Please avoid using too many font style/size. Unify the format: Upper case for the first/all words. }
    % \xiaohan{I will enlarge fonts, still hard to see}
    % \xiaohan{TODO: Maybe do a double column if there is still time.}
    % }
    % \tbd_yan{check}
    }
    \vspace{-1.5em}
\label{fig:overview}
\end{center}
\end{figure}

% Next, we introduce the Large Language Model-Guided Robot Task and Motion Planning (\method{}), which consists of two key components: LLM and Task and Motion Planner, as shown in Fig.~\ref{fig:overview}.
In this paper, we develop \method{} that leverages LLMs to facilitate a mobile manipulator completing object rearrangement tasks. 
\method{} consists of two key components, LLM for generating symbolic spatial relationships (Sec.~\ref{sec:symbolic}) and geometric spatial relationships (Sec.~\ref{sec:geometric}) between objects, and TAMP for computing optimal task-motion plan (Sec.~\ref{sec:plan}), as shown in Fig.~\ref{fig:overview}. 

% The yellow lines represent the adaptive sampling of object positions, where LLM computes the physical distances between objects. 
% To ground the object attributes, we have designed the Object-Grounded Adaptive Sampler to generate a set of feasible object positions.
% Finally, the pink loop demonstrates the selection of optimal object positions with maximal long-term utility for task execution.
% Our proposed \method{} framework is a novel approach that combines the strengths of LLM and Task and Motion Planning to achieve more efficient and effective robot task execution.

% overview link: https://docs.google.com/drawings/d/1y3rNHxSVSuvQll8TV6vJHTm10PbEBSvC43shZ15ZwiM/edit?usp=sharing

\begin{figure*}
% https://docs.google.com/drawings/d/11nxfSwjAcQw8kOdn3jkQrpnTZMxzetWvXQ68k4mV5Ss/edit?usp=sharing
\begin{center}
    \includegraphics[width=0.87\textwidth]{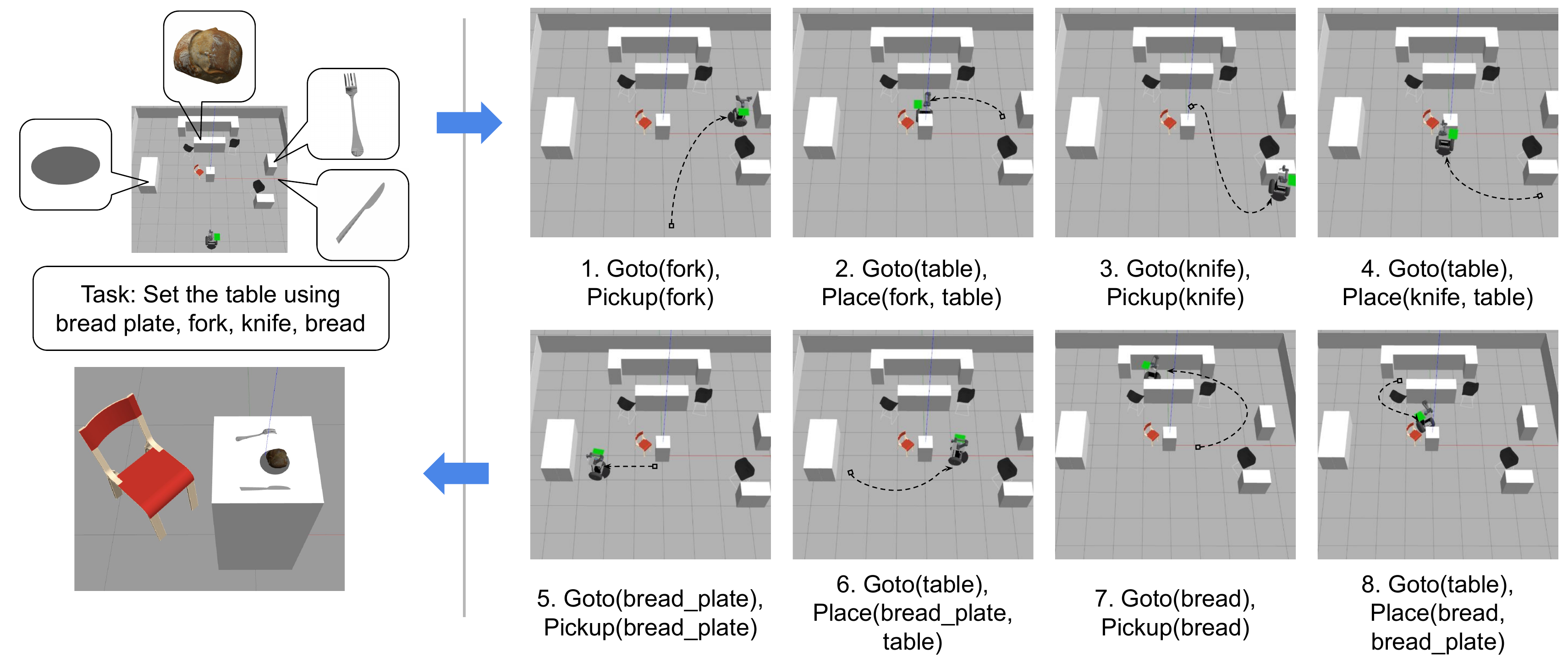}
    % \vspace{-0.5em}
    \caption{An illustrative example of \method{} showing the robot navigation trajectories (dashed lines) as applied to the task of ``set the table with a bread plate, a fork, a knife, and a bread.'' \method{} is able to adapt to complex environments, using commonsense extracted from GPT-3 to generate efficient (i.e., minimize the overall navigation cost) and feasible (i.e., select an available side of the table to unload) pick-and-place motion plans for the robot.
    %Results are shown in the Gazebo simulator. 
    }
    \label{fig:illustrative_example}
    \vspace{-1.5em}
\end{center}
\end{figure*}

\subsection{Generating Symbolic Spatial Relationships}\label{sec:symbolic}
% LLMs are used to extract common sense about the symbolic spatial relationships between objects on a table via the following template-based prompt:
LLMs are used to extract common sense knowledge regarding symbolic spatial relationships among objects placed on a table.
This is accomplished through the utilization of a template-based prompt:

\begin{quote}
% \begin{flushleft}
Template 1: \emph{The goal is to set a dining table with objects. The symbolic spatial relationship between objects includes [spatial relationships]. [examples]. What is a typical way of positioning [objects] on a table? [notes].}
% \end{flushleft}
\end{quote}

\noindent where \emph{[spatial relationships]} includes a few spatial relationships, such as \emph{to the left of} and \emph{on top of}.
%above and to the right of, below and to the right of, below and to the left of, above and to the left of, on, under, and center.''
In presence of \emph{[examples]}, the prompting becomes few-shot; when no examples are provided, it is simplified to zero-shot prompting. 
% determines whether the prompt is Zero-shot or Few-shot. 
In practice, few-shot prompts can ensure that the LLM's response follows a predefined format, though more prompt engineering efforts are needed. 
% , while zero-shot prompts can achieve output diversity.
\emph{[objects]} refers to the objects to be placed on the table, such as \emph{a plate}, \emph{a fork}, and \emph{knife}. 
To control the LLM's output, \emph{[notes]} can be added, such as the example \emph{``Each action should be on a separate line starting with `Place'. The answer cannot include other objects''}.
% In this case, LLM's response might be: ``Place the dinner plate in the center of the table. Place the fork to the left of the dinner plate. Place the knife to the right of the dinner plate.''

LLMs are generally reliable in demonstrating common sense, but there may be times when they produce contradictory results. 
To prevent \textbf{logical errors}, a logical reasoning-based approach has been developed to evaluate the consistency of generated candidates with explicit symbolic constraints.
This approach is implemented on answer set programming (ASP), which is a declarative programming language that expresses a problem as a set of logical rules and constraints~\cite{gebser2008user}. 
In the event of a logical inconsistency, the same template is repeatedly fed to the LLM in an attempt to elicit a different, logically consistent output.
ASP enables recursive reasoning, where rules and constraints can be defined in terms of other rules and constraints, providing a modular approach to problem-solving~\cite{jiang2019task}. 
ASP is particularly useful for determining whether sets of rules and constraints are true or false in a given context.

The approach involves defining spatial relationships, their transitions, and rules for detecting conflicts.
These rules are created by human experts and serve to ensure that the generated context is logical and feasible.
One such rule is \verb|:- below(X,Y),right(X,Y)|, which states that object \verb|X| cannot be both ``below'' and ``to the right of'' object \verb|Y| at the same time.
This rule ensures that the resulting arrangement of objects is physically possible.
An instance of identifying a logical error is provided. 
For example, an LLM may generate instructions for arranging objects as follows:
\begin{enumerate}
    \item {Place fruit bowl in the center of table.}
    \item {\emph{Place {butter knife} above and to the right of {fruit bowl}}.}
    \item {\emph{Place {dinner fork} to the left of {butter knife}}.}
    \item {Place dinner knife to the right of butter knife.}
    \item {\emph{Place {fruit bowl} to the right of {dinner fork}}.}
    % \item \emph{Place salt shaker below and to the right of dinner fork.}
    \item {Place water cup below and to the left of dinner knife.}
    % \item[] \hspace{-1.5em}{7. Place napkin above and to the left of water cup.''}
    % \item \emph{Place salt shaker in the upper left corner of the table..}
\end{enumerate}

There are \textbf{logical inconsistencies} in the italic lines: Steps 2 and 3 suggest placing the \emph{fruit bowl} below the \emph{dinner fork}, while Step 5 suggests placing the \emph{fruit bowl} to the right of the \emph{dinner fork}.
This contradicts the established rule and results in no feasible solutions.
% However, Step 5 implies that ``fruit bowl'' should be placed to the right of ``dinner fork''.
% , according to the transition rule:
% \begin{verbatim}
% below(Y,Z) :- above_right(X,Y),left(Z,X), 
%               not below_right(X,Y), 
%               not below_left(X,Y)
% \end{verbatim}
% However, Step 5 implies that ``fruit bowl'' should be placed to the right of ``dinner fork'', which contradicts the established rule of 
% \verb|:- below(Y,Z),right(Y,Z)|, prohibiting a solution where object \verb|Y| is both ``below'' and ``to the right of'' object \verb|Z| simultaneously.
% This approach helps to make the generation process more reliable and robust by effectively addressing certain types of logical inconsistencies in the generated candidates.
% \begin{enumerate}
%     \item {\emph{Place dinner plate in the center of table.}}
%     \item {\emph{Place {dinner knife} to the right of {dinner plate}}.}
%     \item {\emph{Place {dinner fork} to the left of {dinner plate}}.}
%     \item {\emph{Place cup mat above and to the right of dinner plate.}}
%     \item {\emph{Place teacup on top of {cup mat}}.}
%     % \item \emph{Place salt shaker below and to the right of dinner fork.}
%     % \item[] \hspace{-1.5em}{7. Place napkin above and to the left of water cup.''}
%     % \item \emph{Place salt shaker in the upper left corner of the table..}
% \end{enumerate}

\subsection{Generating Geometric Spatial Relationships}~\label{sec:geometric}
After determining the symbolic spatial relationships between objects in Sec.~\ref{sec:symbolic}, we move on to generate their geometric configurations, where we use the following LLM template.

\begin{quote}
Template 2: \emph{[object A] is placed [spatial relationship] [object B]. How many centimeters [spatial relationship] [object B] should [object A] be placed?}
\end{quote}

For instance, when we use Template 2 to generate prompt ``\emph{A dinner plate is placed to the left of a knife. How many centimeters to the left of the water cup should the bread plate be placed?}'', GPT-3 produces the output ``\emph{Generally, the dinner knife should be placed about 5-7 centimeters to the right of the dinner plate.}''

% To calculate the positions of the objects, we first select one object to be the coordinate origin. 
To determine the positions of objects, we first choose a coordinate origin.
This origin could be an object that has a clear spatial relationship to the tabletop and is located centrally. 
A dinner plate is a good example of such an object.
% \cpax{How do we decide? typically best? this is a bit unclear.}
% \tbd_yan{My approach is to use GPT-3 to suggest an object to place in the center of the table. If GPT-3 cannot provide a suggestion, I will choose the object with the largest area. I rephrased these sentences, what do you think?} 
% It is typically best to choose an object that has a spatial relationship to the table and is located centrally, such as a dinner plate. 
We then use the recommended distances and the spatial relationships between the objects to determine the coordinates of the other objects. 
Specifically, we can calculate the coordinates of an object by adding or subtracting the recommended distances in the horizontal and vertical directions, respectively, from the coordinates of the coordinate origin.
The LLM-guided position for the $i$th object is denoted as $(x^i, y^i)$, where $i\in N$.

However, relying solely on the response of the LLMs is not practical as they do not account for object attributes such as shape and size, including tables constraints.
% \cpax{"including tables" is awk/unclear. You mean table constraints?} 
To address this limitation, we have designed an adaptive sampling-based method that incorporates object attributes after obtaining the recommended object positions.
Specifically, our approach involves sequencing the sampling of each object's position using a 2D Gaussian sampling technique~\cite{boor1999gaussian}, with $(x^i, y^i)$ as the mean vector, and the covariance matrix describing the probability density function's shape.

% \cpax{added a newline - paragraph too long before}
The resulting distribution is an ellipse centered at $(x^i, y^i)$ with the major and minor axes determined by the covariance matrix. 
However, we do not blindly accept all of the sampling results; instead, we apply multiple rules to determine their acceptability, inspired by rejection sampling~\cite{gilks1992adaptive}.
These rules include verifying that the sampled geometric positions adhere to symbolic relationships at a high level, avoiding object overlap, and ensuring that objects remain within the table boundary. 
% For example, we calculate whether each object would fall outside of the table boundary based on their properties (size or shape) under the computed position.
For example, if the bounding box of an object position falls outside the detected table bounds, we reject that sample.
The bounding box of objects and the table are computed based on their respective properties, such as size or shape.
% \cpax{slightly vague again - use bounding box right? you're saying the bounding box falls outside the detected talbe bounds and so we reject that sample.}
% \tbd_yan{I rephrased these sentence. what do you think?}
After multiple rounds of sampling, we can obtain $M$ object configuration sequences.

\subsection{Computing Task-Motion Plans}~\label{sec:plan}
After identifying feasible object configurations on the tabletop in Steps 1 and 2, the next step is to place the objects on the tabletop based on one of object configuration sequences. 
At the task level, the robot must decide the sequence of object placement and how to approach the table. 
For example, if a bread is on top of a plate, the robot must first place the plate and then the bread. 
The robot must also determine how to approach the table, such as from which side of the table.
Once the task plan is determined, the robot must compute 2D navigation goals (denoted as $loc$) at the motion level that connect the task and motion levels. 
Subsequently, the robot plans motion trajectories for navigation and manipulation behaviors. 

% However, completing the task is not easy. 
% The robot must be able to respond to dynamic obstacles, such as chairs around the table, during the rearrangement process. 
% Moreover, not all feasible $loc$ are given equal consideration. 
In the presence of dynamic obstacles, not all navigation goals ($loc$) are equally preferred. 
For instance, it might be preferable for the robot to position itself close to an object for placement rather than standing at a distance and extending its reach. 
A recent approach called GROP~\cite{zhang2022visually} was developed for computing the optimal navigation goal $loc$, which enabled the task-motion plan with the maximal utility for placing each object in terms of feasibility and efficiency given an object configuration $(x^i_j, y^i_j)$, where $0\leq j\leq M$. 
Therefore, for different groups of object configurations, we use GROP to compute the maximal utility value of task-motion plans and select the best one for execution.
Fig.~\ref{fig:illustrative_example} shows one task-motion plan generated using \method{} for a four-object rearrangement task.

\section{Experiments}
In this section, we evaluate the performance of \method{} using the task of rearranging tableware objects. 
The robot needs to compute semantically valid tabletop arrangements, plan to efficiently rearrange the objects, and realize the plan via navigation and manipulation behaviors. 

\vspace{.5em}
\noindent
\textbf{Baselines:}
\method{} is evaluated by comparing its performance to three baselines, where the first baseline is the weakest. 

\begin{itemize}%[leftmargin=.5em]
    \item Task Planning with Random Arrangement (TPRA): 
    This baseline uses a task planner to sequence navigation and manipulation behaviors, while it randomly selects standing positions next to the target table and randomly places objects in no-collision positions on the table. 
    \item LLM-based Arrangement and Task Planning (LATP): It can predict object arrangements using LLMs and perform task planning. It uniformly samples standing positions around the table for manipulating objects. 
    \item GROP~\cite{zhang2022visually}: It considers plan efficiency and feasibility for task-motion planning, and lacks the capability of computing semantically valid arrangements. Similar to TPRA, GROP also randomly places objects in no-collision positions on the table.
\end{itemize}

\vspace{.5em}
\noindent
\textbf{Experimental Setup:} 
A mobile manipulator is assigned the task of setting a dinner table using a specific set of objects. 
In a simulated environment\footnote{Implemented in the Gazebo simulator}, the robot needs to retrieve multiple objects from various locations and place them on the central table. 
Additionally, an obstacle (i.e., a chair) will be randomly placed around the table. There are eight tasks that involve handling different objects, as detailed in TABLE~\ref{table:task}. 
We execute each task 20 times using the \method{} system with the same prompt templates, and after each task is completed, we capture an image of the table, the chair, and the objects on the tabletop for later human evaluation.
To carry out our experiments, we used OpenAI's GPT-3 engines.
Please refer to TABLE~\ref{table:parameter} for the specific hyperparameters we adopted.
% We opted for the most capable and expensive engine, ``text-davinci-003,'' and set the GPT-3 ``temperature'' to 0.0 to minimize randomness in responses. We also set the ``top\_p'' parameter to 1.0 to increase response diversity. Other parameters such as ``maximum length,'' ``presence\_penalty,'' and ``frequency\_penalty,'' were set to 512.0, 0.0, and 0.0, respectively.
We have chosen not to use ChatGPT, a well-known language LLM, for large-scale experiments due to the unavailability of its APIs.
% In Step 1 of \method{}, the symbolic spatial relationships between two objects in two-dimensional space include ``above'', ``to the right of'', ``below'', ``to the left of'', ``above and to the right of'', ``below, and to the right of'', ``below and to the left of'', and ``above and to the left of''.
% We also consider the ``on top of'' and ``under'' relationships in the vertical direction.
% The mobile manipulator includes a UR5e robot arm, a Robotiq 2F-140 gripper, an RMP 110 mobile base, and a Velodyne VLP-16 lidar sensor. 
% We used a Rapidly exploring Random Tree (RRT) approach~\cite{lavalle1998rapidly} to compute motion-level manipulation plans. 
% The navigation stack was built using the \texttt{move\_base} package of Robot Operating System (ROS)~\cite{quigley2009ros}. 
% The robot's task planner is  ASP-based~\cite{lifschitz2002answer} and we used the Clingo solver for computing task plans~\cite{gebser2014clingo}.

\begin{table}[t]
\scriptsize
% \footnotesize
\centering
\caption{Objects that are involved in our object rearrangement tasks for evaluation, where tasks 1-5 include three objects, tasks 6 and 7 include four objects, and task 8 includes five objects.}\label{table:task}
\begin{tabular}[t]{@{}cl@{}}
\toprule
	 Task \#ID & Objects\\ \midrule
	 1 & Dinner Plate, Dinner Fork, Dinner Knife\\ \midrule
      2 & Bread Plate, Water Cup, Bread\\ \midrule
	 3 & Mug, Bread Plate, Mug Mat\\ \midrule
      4 & Fruit Bowl, Mug, Strawberry\\ \midrule
      5 & Mug, Dinner plate, Mug Lid\\ \midrule\midrule
      6 & Dinner Plate, Dinner Fork, Mug, Mug Lid\\ \midrule
      7 & Dinner Plate, Dinner Fork, Dinner Knife, Strawberry\\ \midrule\midrule
      8 & Dinner Plate, Dinner Fork, Dinner Knife, Mug, Mug Lid\\
\bottomrule
\end{tabular}
\vspace{-0.5em}
\end{table}

\begin{table}[t]
\caption{Hypermeters of OpenAI's GPT-3 engines in Our Experiment}\label{table:parameter}
% \footnotesize
\scriptsize
\centering
\begin{tabular}{l|l||l|l}
% \begin{tabular}{|p{2cm}|p{2cm}|p{2cm}|p{2cm}|}
\toprule
\textbf{Parameter} & \textbf{Value} & \textbf{Parameter} & \textbf{Value} \\
\midrule
Model & text-davinci-003 & Temperature & 0.1 \\
\midrule
Top p & 1.0 & Maximum length & 512 \\
\midrule
Frequency penalty & 0.0 & Presence penalty & 0.0 \\
\bottomrule
\end{tabular}
% \vspace{-1.em}
\end{table}

\begin{table}[!t]
\caption{Rating guidelines for human raters in the experiments. \textbf{1} point indicates the poorest tableware object arrangement as it suggests that some objects are missing. Conversely, \textbf{5} points represent the best arrangement.}
\label{table:criteria}
\begin{threeparttable}
\scriptsize
% \footnotesize
\begin{tabular}{cl}
\toprule
	 Points & Rating Guidelines\\ \midrule
	 1 & Missing critical items compared with the objects listed at the top of \\
        & the interface (e.g., dinner plate, dinner fork, dinner knife), making it \\
        & hardly possible to complete a meal.\\ \midrule
      2 & All items are present, but the arrangement is poor and major \\
        & adjustments are needed to improve the quality to a satisfactory level.\\ \midrule
	 3 & All items are present and arranged fairly well, but still there is\\
        & significant room to improve its quality.\\ \midrule
      4 & All items are present and arranged neatly, though an experienced\\
        & human waiter might want to make minor adjustments to improve.\\ \midrule
      5 & All items are present and arranged very neatly, meeting the aesthetic \\
        & standards of an experienced human waiter.\\
\bottomrule
\end{tabular}
\end{threeparttable}
\vspace{-1.5em}
\end{table}

\vspace{.5em}
\noindent
\textbf{Rating Criteria:}
% To evaluate the quality of tableware object arrangements generated by \method{} and three baselines, 
We recruited five graduate students with engineering backgrounds, three females and two males between the ages of 22 and 30. 
We designed a five-point rating rule, which is outlined in Table~\ref{table:criteria}, and tasked the volunteers with scoring tableware object rearrangements in images they were shown.
We generated 640 images from the four methods (three baselines and LMM-GROP) for eight tasks and each image required evaluation from all volunteers, resulting in a total sample size of 3200 images. 
The volunteers were shown one image at a time on a website\footnote{The link for the questionnaire-based experiment results evaluation is \url{http://150.158.148.22/}} that we provided, and they scored each image from 1 to 5 based on the rating rules.
We ensured that the rating was rigorous by using a website to collect rating results, thereby minimizing any potential biases that could arise from further interaction with the volunteers once they entered the website. 
%For further information about our methodology, please refer to our project website\footnote{\textcolor{red}{Supplementary materials are available at: \url{https://sites.google.com/view/llm-grop}}}.

\begin{figure}
% https://docs.google.com/drawings/d/11nxfSwjAcQw8kOdn3jkQrpnTZMxzetWvXQ68k4mV5Ss/edit?usp=sharing
\begin{center}
\vspace{-0.5em}
    \includegraphics[width=0.45\textwidth]{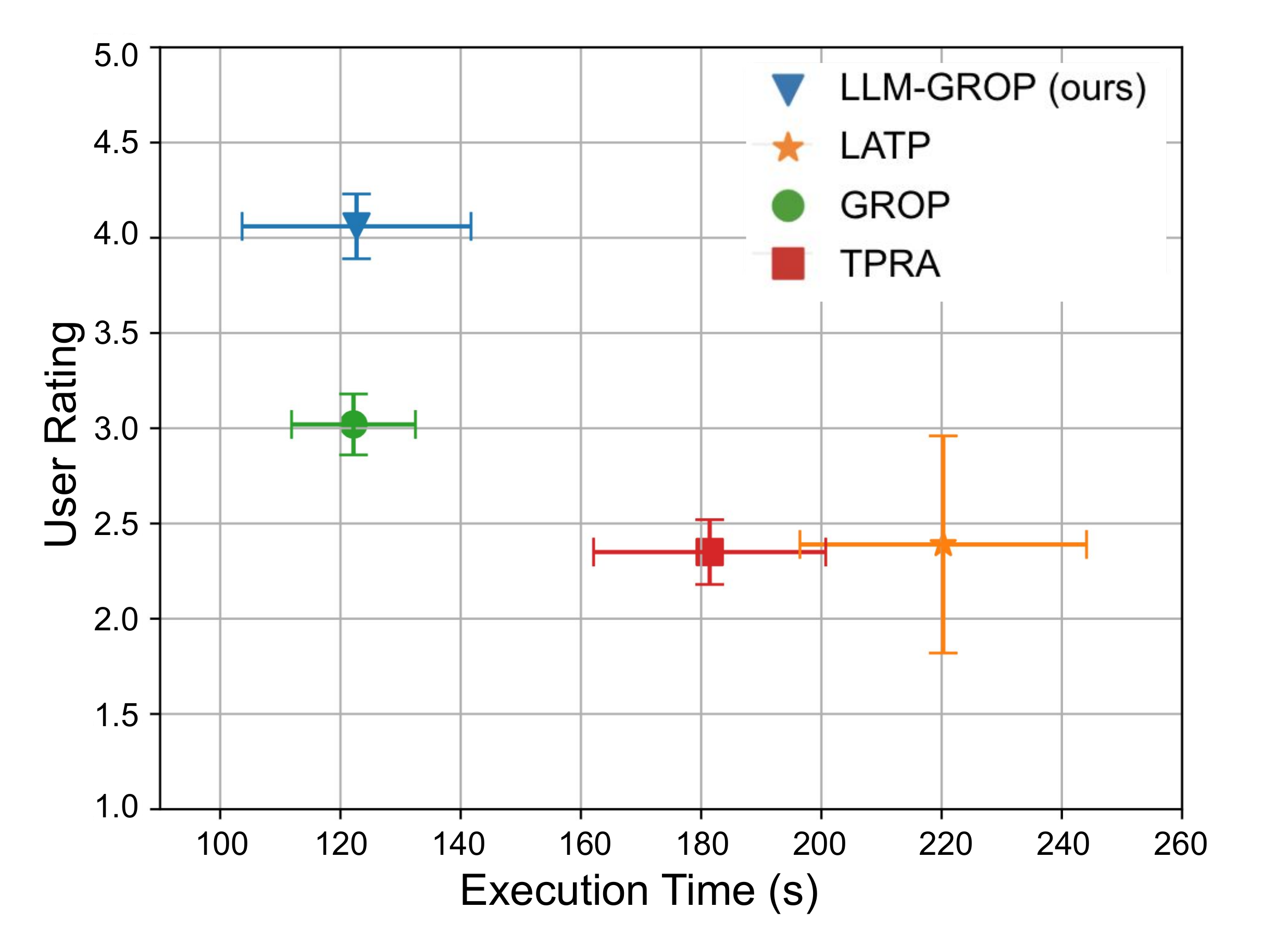}
    \caption{Overall performance of \method{} as compared to three baselines based on mean values and standard errors of user ratings and robot execution time for all tableware object arrangement tasks. %\shiqi{Are you reporting standard errors or standard deviations using the bars? I think it's the former given that they are so small. Also, baseline names. }
    }\label{fig:main}
    \vspace{-1.em}
\end{center}
\end{figure}

\vspace{.5em}
\noindent
\textbf{\method{} vs. Baselines:} 
Fig.~\ref{fig:main} shows the key findings of our experiments, which compares the performance of \method{} to the three other baseline approaches. 
The $x$-axis indicates the time each method takes to complete a single task, while the $y$-axis indicates the corresponding user rating.
The results demonstrate that our \method{} achieves the highest user rating and the shortest execution time compared to the other approaches. 
While GROP proves to be as efficient as our approach, it receives a significantly lower rating score. 
By contrast, \random{} and \LLM{} both receive lower user ratings than our \method{}.
They also display poor efficiency.
This is because they lack the navigation capabilities to efficiently navigate through complex environments.
For instance, when their navigation goals are located within an obstacle area, they struggle to adjust their trajectory, leading to longer task completion times.

\begin{figure}
% https://docs.google.com/drawings/d/152wBq8Sh9pZOc9jfHRiru5YRMHmIGuwObhjcDHBkT4U/edit?usp=sharing
\begin{center}
    \includegraphics[width=0.48\textwidth]{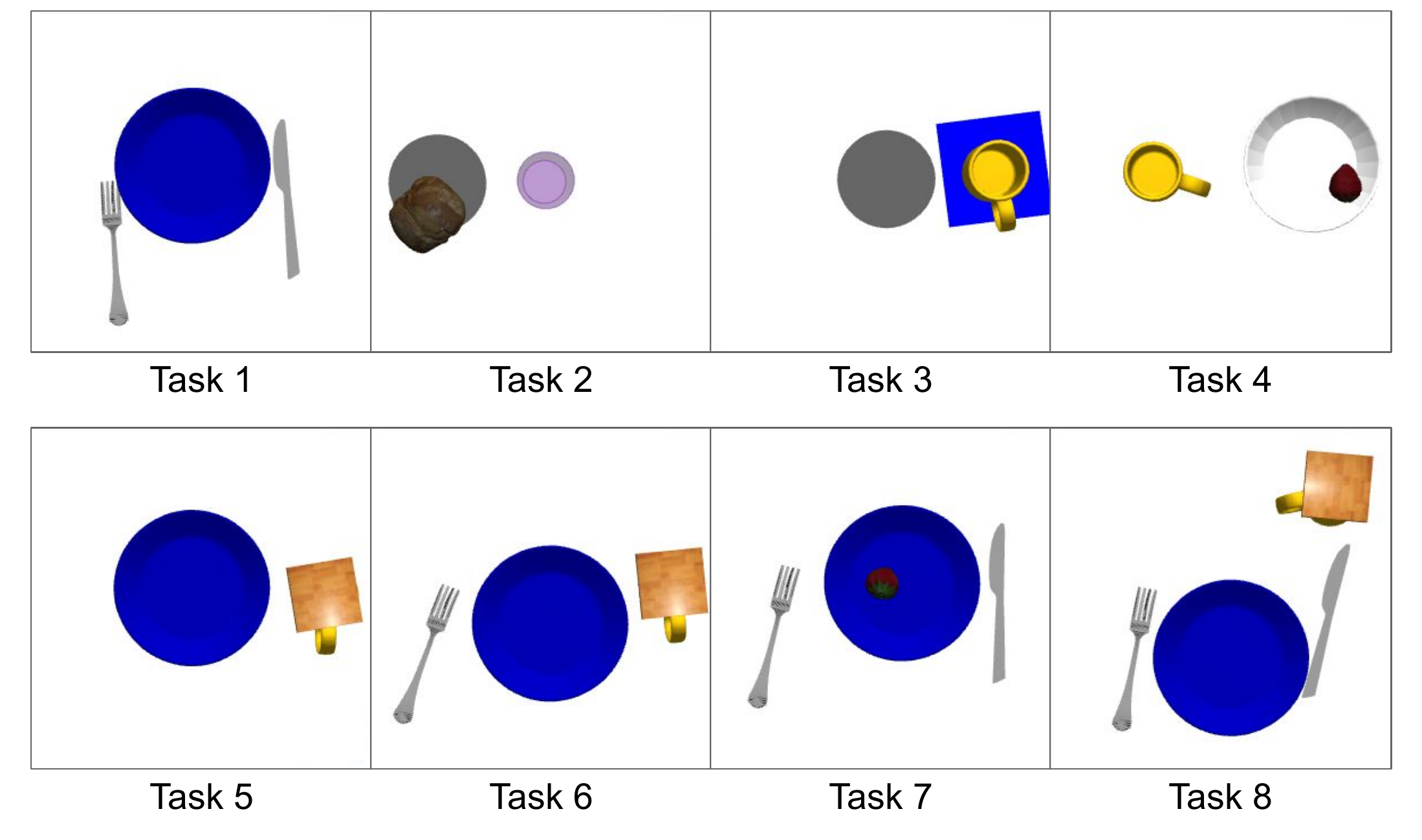}
    \vspace{-1.em}
    \caption{Examples of tableware objects rearranged by our LLM-GROP agent in eight tasks, where the objects used in these tasks can be found in Table~\ref{table:task}.
    % To better suits user preferences, 
    Our \method{} enables the arrangement of tableware objects to be both semantically valid.
    %Our method can reason about symbolic spatial relationships, such as left, on top of, above to the right, and more, to ensure semantically valid and functional outcomes that suit the specific environment.
    % To compute a feasible plan for the robot to execute, our method can fine-tune geometric spatial relationships to adapt to the complex environment while ensuring to satisfy symbolic constraints.
    }\label{fig:8examples}
\end{center}
\vspace{-1.5em}
\end{figure}

\begin{figure*}
% https://docs.google.com/drawings/d/11nxfSwjAcQw8kOdn3jkQrpnTZMxzetWvXQ68k4mV5Ss/edit?usp=sharing
\begin{center}
    \includegraphics[width=0.97\textwidth]{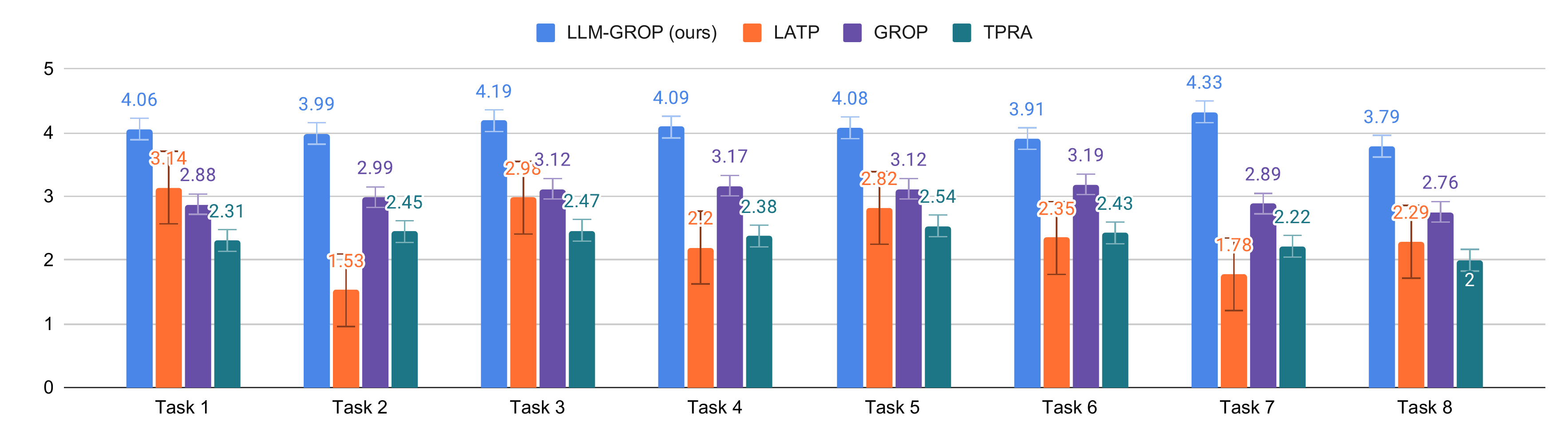}
    \vspace{-0.85em}
    \caption{User ratings of individual object rearrangement tasks, with the $x$-axis representing the task and the $y$-axis representing the user rating score. It can be observed that \method{} consistently performs the best compared to baselines. Tasks 1-5 involve three objects, tasks 6 and 7 involve four objects, and task 8 involves five objects. The numerical value displayed on each bar indicates the mean rating for the corresponding task.
    }\label{fig:individual}
\end{center}
% \vspace{-1.5em}
\end{figure*}

%(old)https://docs.google.com/drawings/d/1AiMkGVzBBidSYDZ428dyvt_ma5yOo5dfWrua83mCFd4/edit?usp=sharing
%(new)https://docs.google.com/drawings/d/1aB8c8qtS8B8j8gNUvdnqjXgkxEBQQcrY-WjIrzgT8oU/edit?usp=sharing
\begin{figure*}
\begin{center}
\includegraphics[width=0.98\textwidth]  {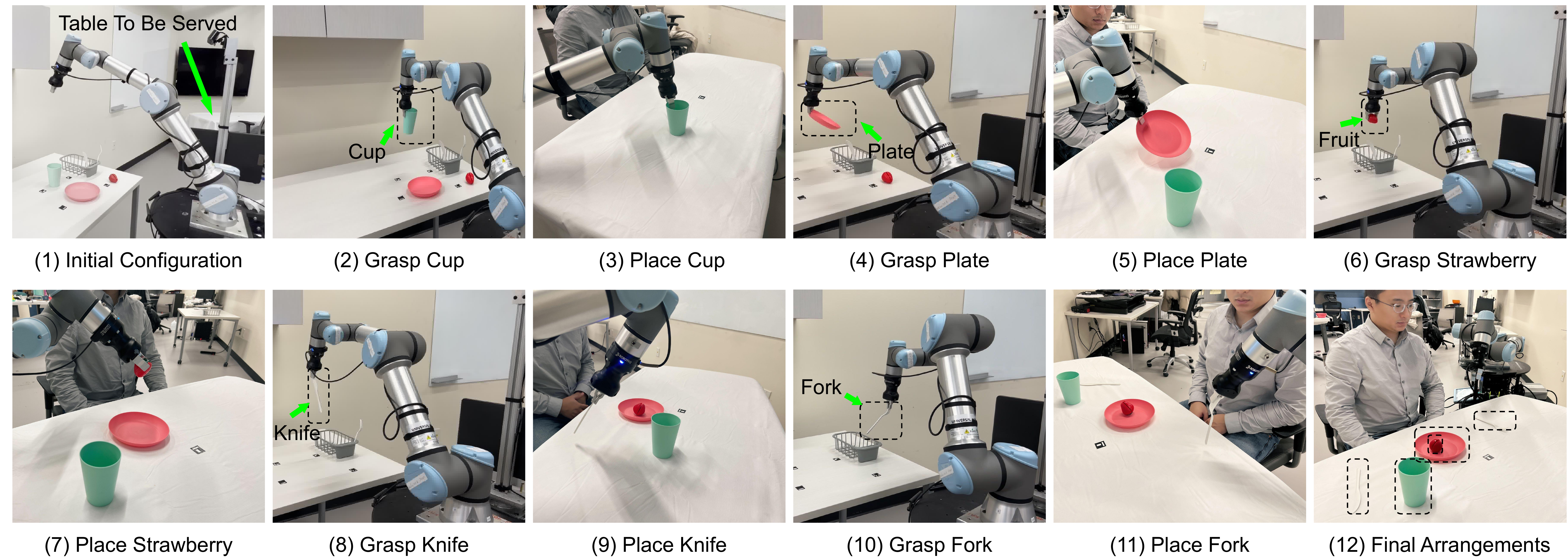}
    % \vspace{-1.em}
    \caption{We demonstrate LLM-GROP on real robot hardware.
    The real-robot system includes a Segway-based mobile platform and a UR5e robot arm.
    The robot employs hard-coded procedures for object grasping.
    The task is to serve a human with a knife, a fork, a cup, a plate, and a strawberry.
    The robot computes a plan that successfully avoids chairs and the human around the table, while being able to place the target objects in plausible physical positions. 
    % \textcolor{red}{SHIQI: One can hardly tell what the target table is like. To show the initial and final configurations, it's better to have top-down views. To show the two tables and/or the robot, it's better to use a 3rd person view. }
    % \textcolor{blue}{Yan: If you think it's absolutely essential to update the image, I can take a new photo.}
    % \shiqi{What do you mean by ``Right Unloading'' or ``Left Unloading"? }
    % \yan{"loading objects in different standing positions", which shows the capability of \method{} in selecting the optimal standing position. Looks like these two phrases make readers confusing. I have removed them and mentioned "loading objects in different standing positions" in the caption.}
    }\label{fig:realrobot}
\end{center}
\end{figure*}

Fig.~\ref{fig:8examples} provides several examples of various tasks that are rearranged by our agent.
Fig.~\ref{fig:individual} presents the individual comparison results of each method for individual tasks. 
The $x$-axis corresponds to Task \#ID in Table~\ref{table:task}, while the $y$-axis represents the average user rating for each method.
Our \method{} demonstrates superior performance over the baselines for each task. 
Specifically, tasks 1 to 5 receive slightly higher scores than tasks 6 and 8. 
This is reasonable because the latter two tasks require the robot to manipulate more objects, posing additional challenges for the robot.
% \shiqi{Fig. 3 should be move closer to this paragraph, and the figures should appear in the order of where they are referred to. }

\vspace{.5em}
\noindent 
\textbf{Real Robot Demonstration: }
We tested our \method{} approach on a real mobile robot platform to demonstrate its effectiveness in rearranging a set of tableware objects, as shown in Fig.~\ref{fig:realrobot}.
The set included a dinner plate, a dinner fork, a dinner knife, a water cup, and a strawberry.
The robot started on the left table and is tasked with rearranging the objects on the right table in the left image.
After successfully completing the task, the robot successfully rearranged the objects as shown in the right image. 
The final object placements were semantically valid, such as the fork being on the left of the dinner plate and the strawberry being on the plate. 
These outcomes effectively demonstrate the effectiveness of our approach in performing real-world tasks using a robotic platform.
We have generated a demo video that has been uploaded as part of the supplementary materials.

\section{Conclusion and Future Work}
To summarize, we propose \method{}, which demonstrates how we can extract semantic information from LLMs and use it as a way to make commonsense, semantically valid decisions about object placements as a part of a task and motion planner - letting us execute multi-step tasks in complex environments in response to natural-language commands. 
In the future, we may take more information from methods like M0M~\cite{curtis2022long}, in order to perform grasping and manipulation of fully unknown objects in unknown scenes, and expand to a wider set of placement problems.

\section*{ACKNOWLEDGMENTS}
A portion of this work has taken place at the Autonomous Intelligent Robotics (AIR) Group, SUNY Binghamton. 
AIR research is supported in part by grants from the National Science Foundation (NRI-1925044), Ford Motor Company (URP Award 2019-2023), OPPO (Faculty Research Award 2020), and SUNY Research Foundation.

% The first paragraph of Section III of the manuscript
% briefly talks about the consideration of uncertainty in
% navigation and manipulation behaviors, however no further
% details are provided in the manuscript of how exactly
% uncertainty is considered in the proposed approach. Does
% LLM-GROP take a probabilistic inference approach to account
% for uncertainty? Is a replanning routine instead triggered
% whenever the robot fails to navigate to a goal or grasp an
% object? Further details on the strategy for accounting
% uncertainty would certainly improve the manuscript.

%     \item some details about the real
% robot configuration and the tests performed highlighting
% any possible difference from the simulated one will be of
% interest. 

\bibliographystyle{IEEEtran}  

\bibliography{references}

\end{document}